\documentclass[letterpaper, 10 pt, conference]{ieeeconf}  

\IEEEoverridecommandlockouts                              

\usepackage{graphics} 
\usepackage{epsfig} 
\usepackage{mathptmx} 
\usepackage{times} 
\usepackage{amsmath} 
\usepackage{amssymb}  
\usepackage{booktabs}
\usepackage{graphicx}   
\usepackage{caption}
\usepackage{subcaption}

\title{\LARGE \bf
Integrating Trajectory Optimization and Reinforcement Learning for Quadrupedal Jumping with Terrain-Adaptive Landing}

\author{Renjie Wang, Shangke Lyu$^{*}$, Xin Lang, Wei Xiao, Donglin Wang$^{*}$
\thanks{This work was supported by the National Natural Science Foundation of China (Grant No. 62176215 and 62003018).}
\thanks{All the authors are affilated with Machine Intelligence Lab (MiLAB), School of Engineering, Westlake University, Hangzhou 310024, China.}
\thanks{$^{*}$Corresponding author.}
}

\begin{document}

\maketitle
\thispagestyle{empty}
\pagestyle{empty}

\begin{abstract}
Jumping constitutes an essential component of quadruped robots' locomotion capabilities, which includes dynamic take-off and adaptive landing. Existing quadrupedal jumping studies mainly focused on the stance and flight phase by assuming a flat landing ground, which is impractical in many real world cases. This work proposes a safe landing framework that achieves adaptive landing on rough terrains by combining Trajectory Optimization (TO) and Reinforcement Learning (RL) together. The RL agent learns to track the reference motion generated by TO in the environments with rough terrains. To enable the learning of compliant landing skills on challenging terrains, a reward relaxation strategy is synthesized to encourage exploration during landing recovery period. Extensive experiments validate the accurate tracking and safe landing skills benefiting from our proposed method in various scenarios.
\end{abstract}

\section{INTRODUCTION}
With the development of the legged robot community over several decades, the locomotion performance of quadruped robots has improved remarkably, including but not limited to traversing the wild and performing robustly against disturbances~\cite{Grandia_2023,Lee_2020,Miki_2022,lyu2023composite,Jenelten_2024,margolis2022walk,nahrendra2023dreamwaq,lyu2024rl2ac,lyu2025robotic}. However, very limited works have demonstrated the ability to robustly jump onto challenging and discontinuous terrains with safe landing. This ability is inherently exhibited by their legged animal counterparts. For example, goats are known for their ability to navigate steep, rocky surfaces, jumping with accuracy and landing securely, thanks to their adaptive limb coordination and balance.

To achieve agile and dynamic quadrupedal jumping, model-based Trajectory Optimization (TO) is one of the preferred methods. In such methods, a simplified dynamics model is commonly required to ease the optimization process and there are several different ways to derive simplified equivalent dynamics~\cite{Nguyen, Kang_VHIMP, Ding_RAL}.
Among them, the single rigid body model (lumped mass model) benefits from considering centroidal dynamics, which ignores the dynamic influence of individual links~\cite{chignoli2022}. This enables the generation of complex motions with a relatively low computation cost, making it suitable for producing omnidirectional 3-D jumping motion. 
For structures of TO problems, kino-dynamic TO highlights its planning stability by avoiding joint collisions or exceeding kinematic limits. This is achieved by adding kinematic constraints that enforce the mapping between the joint and the center of mass (CoM) states, and by jointly optimizing the motions in both spaces~\cite{Ding_impactaware}. However, this model-based method may be seriously unreliable when robots locomote in unstructured environments like rough terrains where simplifying assumptions are violated, which is still an open research area.

Reinforcement learning (RL) has been extensively demonstrated as a powerful paradigm for legged locomotion. Unlike model-based methods, RL agents directly learn control behavior through trial-and-error~\cite{Hwangbo_2019}. Millions of interactions during training enable robustness against external uncertainties and disturbances, and domain randomization mitigates sim-to-real gap~\cite{Lee_2020,Miki_2022}. These factors make RL policies perform impressively in corner cases where modeling changes in the environment and disturbances is difficult for TO. However, most RL frameworks for challenging environments and highly dynamic motions like jumping suffer from inefficient sampling during the training process, inaccurate motion compared to model-based methods and tedious reward shaping.     
These problems can be mitigated by motion imitation, in which a policy is trained to replicate reference motion obtained from real-world demonstrations or other controllers. Cheng et al.~\cite{cheng2024express} proposed a framework to learn and imitate human expressive behaviors for humanoid robots by rewarding joint space tracking of human motion capture data. Moreover, Jenelten et al.~\cite{Jenelten_2024} and Kang et al.~\cite{kang_MPCRL} demonstrated similar motion tracking on quadruped robots, except that the prior knowledge of motion was rolled out from Model Predictive Control (MPC). For quadrupedal jumping, existing work that trains policies for this motion is rare. Bellegarda et al.~\cite{bellegarda2024} learned a policy that rectifies the tracking error of reference motion produced by TO. Controlling jump motion using an RL policy directly has been demonstrated by Atanassov et al.~\cite{atanassov_2024} using curriculum learning, although reward shaping and tuning are tedious in this work.

\begin{figure*}[t]
  \centering
  \includegraphics[width=1\textwidth]{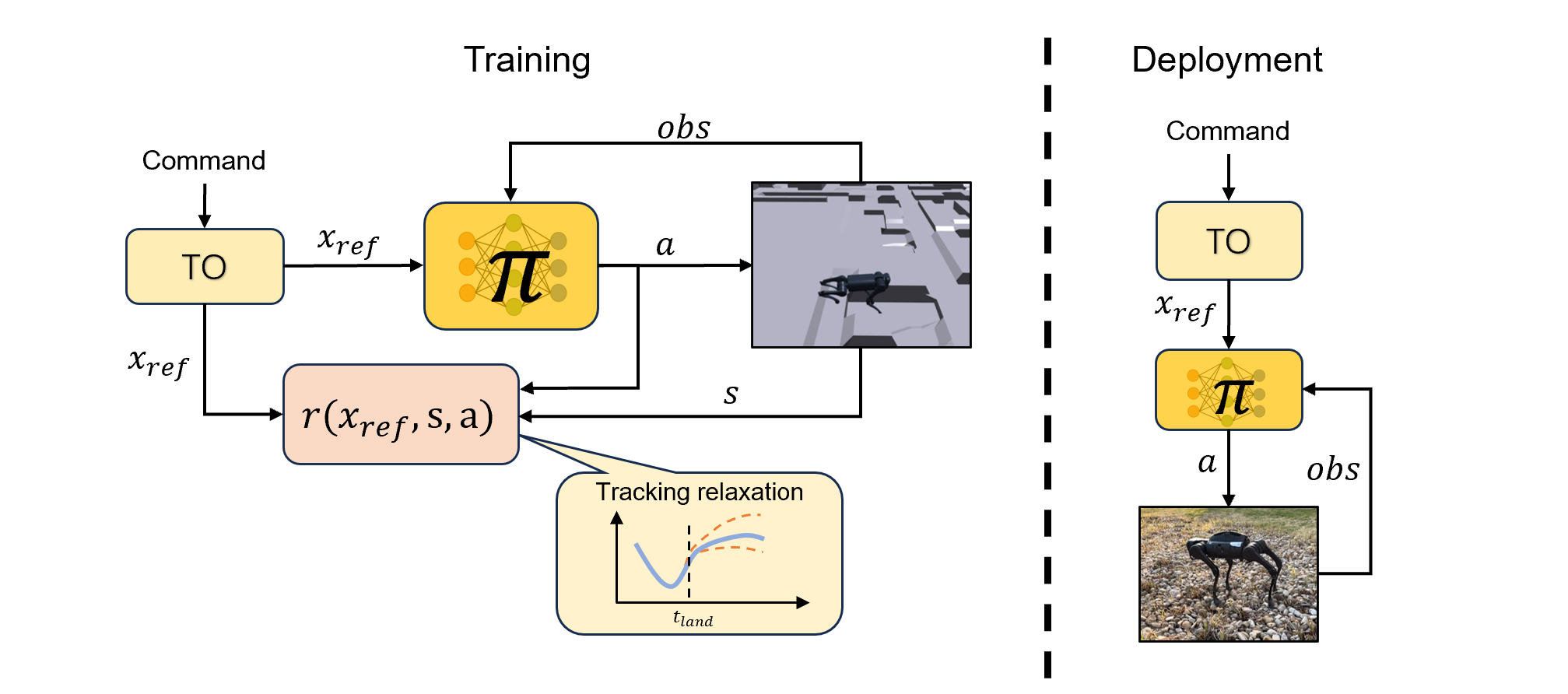} 
  \caption{Overview of the proposed method}
  \label{fig:framework}
\end{figure*}

The above studies, both using model-based methods and RL, mostly focused on landing on flat ground or terrains with mild obstacles, while in reality landing terrains are not in this case, where extreme landing terrains may be introduced by challenging environments or a failed execution of jumping motion (e.g., getting trapped when jumping over a high obstacle). Although there are some works that studied adaptive landing under aggressive horizontal velocities or body poses, they are only applicable on flat ground~\cite{Ding_impactaware,roscia2023,Nguyen_2022}. Existing studies of compliant landing on uneven terrains are limited to low-gravity environments~\cite{Rudin_2022,QI2023}, while related works under earth gravity are limited. In this work, we propose that safe landing is equally important for quadrupedal jumping task. This work thus aims to demonstrate such a control architecture with experimental validation on hardware. More specifically, to achieve successful blind landing on challenging terrains, we observe that the robot should 1) perform dynamic take-off against disturbances and 2) land on unknown terrains adaptively. It is difficult to use TO to model such unstructured landing terrains, while RL can be leveraged to learn such a capability from data. In order to simplify reward shaping and tuning for the stance phase and flight phase, we leverage the reference trajectories solved from TO in RL with a motion tracking framework. To overcome the limitation of reference joint trajectories that become suboptimal on rough terrains, a reward relaxation strategy is synthesized to encourage exploration for an adaptive landing behavior. To the authors' best knowledge, this is the first time a quadrupedal jumping and safe landing controller has been demonstrated on rough terrains.

To this end, we summarize the main contributions of this work as follows:
\begin{itemize}
\item An effective and efficient training framework for dynamic jumping by learning to imitate reference motion rolled out from TO, which eliminates tedious reward shaping and curriculum design.
\item A relaxation strategy for the joint tracking reward during the landing phase, which enables the compliant joint behavior to adapt to uneven terrains.
\item Extensive hardware experiments that validate both the accurate joint tracking ability and the remarkable compliant joint skill, which improve the survival rate of landing in various scenarios including rough terrains, slippery ground, and unknown disturbances.    
\end{itemize}

\section{Method}
In this section we describe our dynamic jumping and adaptive landing control framework that combines TO and RL, as well as the reward relaxation that enables the acquisition of a compliant landing skill, as depicted in Fig.~\ref{fig:framework}.

\subsection{Generating Jumping Motion Behaviors}
We use a modified version of the kino-dynamic trajectory optimization for omnidirectional jump proposed by Ding et al.~\cite{Ding_impactaware}.
\subsubsection{Dynamics model}
The TO takes a centroidal dynamics model, and the motion is govened by

\begin{subequations}\label{eq:centroid_dyn}
\begin{equation}\label{eq:stance_dyn}
\begin{bmatrix}
\ddot{\mathbf{r}} \\ 
\dot{\mathbf{h}}
\end{bmatrix}
= 
\begin{bmatrix}
\displaystyle \frac{1}{m} \sum_{i=1}^{{n}_{\text{c}}}  \mathbf{f}_{i} + \mathbf{g} \\ 
\displaystyle \sum_{i=1}^{n_{\text{c}}} (\mathbf{c}_i - \mathbf{r})\times \mathbf{f}_{i}
\end{bmatrix}
\end{equation}
\begin{equation}\label{eq:flight_dyn}
\ddot{\mathbf{r}} = \mathbf{g}    
\end{equation}
\end{subequations}
for stance and flight phase respectively, where \(\mathbf{r} \in \mathbb{R}^3\), \(\mathbf{h} \in \mathbb{R}^3\), \(\mathbf{f}_i\in \mathbb{R}^3\), \(\mathbf{c}_i\in \mathbb{R}^3\), \(n_{\text{c}}\), \(\mathbf{g}\in \mathbb{R}^3\)and \(m\) respectively denote the 3-D CoM position, centroid angular momentum, ground reaction force (GRF), contact location, number of feet in contact, gravitational acceleration and mass of the robot.

\subsubsection{Kino-dynamic jump motion optimization}
The optimization assumes that the four feet lift off simultaneously and the body remains its homing pose during the whole jump period with a constant \(\mathbf{h}\).
Under the second assumption, the distribution of GRF at each foot along the trajectories can be derived from the second row of Equation~\eqref{eq:stance_dyn} and the predefined feet polygon, as detailed in \cite{Ding_impactaware}. 
Thanks to the coupled TO, the CoM trajectories and joint trajectories can be solved from a single optimization problem. At a high level, the optimization problem can be formulated as follows:
\begin{subequations}\label{eq:TO_formula}
\begin{align}
&  \min_{\mathbf{X}, \mathbf{Q}, \mathbf{U}, \mathbf{t}, \xi} \quad J_{\text{cost}}(\mathbf{X}, \mathbf{Q}, \mathbf{U}, \mathbf{t}, \xi) \label{eq:TO_cost} \\
\text{s.t.} \quad & \mathbf{X}_{k+1} = f_{\text{EoM}}(\mathbf{X}_k,\mathbf{U}_k,\mathbf{t}), \quad k = 1 \ldots N - 1 \label{eq:TO_dyn_cons} \\
& \mathbf{X}_k = \text{kinematics}(\mathbf{Q}_k), \quad k = 1 \ldots N_s-1 \label{eq:TO_kin_cons} \\
& \mathbf{X}_{\mathbf{k}} \in \mathbb{X}_{\text{s}} , \quad \mathbf{k} \in \mathbb{K}_{\text{s}} \label{eq:TO_specific_cons} \\
& \mathbf{X} \in \mathbb{X}, \quad \mathbf{Q} \in \mathbb{Q}, \quad \text{torque}(\mathbf{X},\mathbf{Q},\mathbf{U}) \in \mathbb{T} \label{eq:TO_bounds}
\end{align}
\end{subequations}
where \(\mathbf{X} \in \mathbb{R}^{6 \times (N_s+N_f)}\) is a set of body CoM position and velocity trajectories, \(\mathbf{Q} \in \mathbb{R}^{12 \times N_s}\) and \(\mathbf{U} \in \mathbb{R}^{3n_{\text{c}} \times N_s}\) are, respectively, the trajectories of joint positions and contact forces in stance phase, \(\mathbf{t} \in \mathbb{R}^2\) contains the timesteps for stance phase and flight phase, and \(\xi\) denotes a slack variable for constraint relaxation. To formulate the optimization, the jumping trajectories are discretized into a set of \(N_s\) timesteps regarding to stance phase and a set of \(N_f\) timesteps for flight phase, respectively, thus the total timesteps \(N = N_s + N_f\). 

The objective function \(J_\text{cost}\) consists of several terms, and each term is corresponding to the cost of a specific phase or phase transition moment: 
\begin{equation}\label{eq:cost_details}
J_{\text{cost}} = J_{\text{stance}} + J_{\text{takeoff}} + J_{\text{flight}} + J_{\text{land}} + J_{t} + J_{q} + J_{\xi},    
\end{equation}
where the first four subterms correspond to the synthesis of different motion phases, i.e., stance phase, take-off, flight phase and landing respectively, and the last three denote the regularization of timesteps, joint angles and slack variable respectively.
The solution of the optimization problem must satisfy a set of constraints that can be categorized in the following way:
\begin{itemize}
\item Equation~\eqref{eq:TO_dyn_cons} denotes the dynamics constraints for the whole jump motion, where \(f_{\text{EoM}}\) defines the discretized equation of motion (EoM).
\item Equation~\eqref{eq:TO_kin_cons} denotes the kinematics constraints over the stance period.
\item Equation~\eqref{eq:TO_specific_cons} includes single-knot constraints that restrict the robot's states at fixed positions or limit them within a predefined range at specific moments, i.e. initial state, take-off and landing moment.
\item Equation~\eqref{eq:TO_bounds} contains ranges of the decision variables. Note that the contact forces \(\mathbf{U}\) are constrained by torque limits in joint space using contact Jacobian. 
\end{itemize}
We refer the the work of~\cite{Ding_impactaware} for the detailed formulation and will only elaborate our changes.


A complete jumping and landing planner was proposed in~\cite{Ding_impactaware} that includes kino-dynamic jumping motion generation, and leg motion modulation (LMM) and landing recover (LR) for landing adaptation. We only use the planner for jumping motion generation, since the LMM and LR module are under the assumption of a flat landing ground, which would become unreliable for uneven terrains. Instead of tracking the optimized trajectories after touch-down, we intend to let the agent learn such an adaptive landing ability despite the motion reference given to the policy is based on flat ground, which will be detailed in Section~\ref{sec:2-3}

\subsection{Combining TO and RL} \label{sec:2-2}
From a high-level perspective, our framework cascades TO and RL, where the output of TO sequentially feeds into the RL module. Then, we train control policies for quadruped robots to track the reference jumping motion with Proximal Policy Optimization (PPO)~\cite{schulman2017PPO}.
\subsubsection{Observation and Action Space}
The observation of the jumping policy comprises proprioceptive measurements such as  base angular velocity, gravity vector, joint positions and joint velocities. The history only includes previous action. Additional observations are from jumping motion planner, including desired joint positions and a jumping triggering binary. The action space contains joint target positions, which will be sent to a joint PD controller that produces torque for each joint. 
\subsubsection{Reward Definition}
The total reward is defined as the weighted combination of several components, which can be specified as two categories: tracking of the reference motion \(r_{\textit{T}}\) and regularization terms \(r_{\textit{R}}\), namely
\begin{equation}\label{eq:reward_def}
r = r_{\textit{T}} + r_{\textit{R}}.    
\end{equation}
The motion tracking terms encourage the imitation of reference trajectories, which can be defined as
\begin{equation}\label{eq:reward_track}
r_{\textit{T}} = r_{\textit{B}} + r_{\textit{J}},    
\end{equation}
where \(r_{\textit{B}}\) rewards tracking of base pose and \(r_{\textit{J}}\) rewards tracking of joint positions. Both \(r_{\textit{B}}\) and \(r_{\textit{J}}\) can be formulated as
\begin{equation}\label{eq:tracking_reward}
r_{x} = \exp \left( - \left\| \frac{x^*(t) \ominus x(t)}{\sigma_{x}} \right\|^2 \right),    
\end{equation}
where \(x(t)\) is the measured base pose or joint position, \(x^*(t)\) is the desired base pose or joint position sampled from reference trajectories, and \( \ominus \) denotes the quaternion difference for base orientation and vector difference otherwise.
The regularization term is define as
\begin{equation}\label{eq:reward_regu}
r_{\textit{R}} = r_{\Delta a} + r_{\textit{smooth}} + r_{\textit{energy}} + r_{\textit{torque}} + r_{\textit{bc}},    
\end{equation}
where \(r_{\Delta a}\) and \(r_{\textit{smooth}}\) regulate rate of change and smoothness of actions respectively, \(r_{\textit{energy}}\) penalizes joint power to save energy, \(r_{\textit{torque}}\) minimizes joint torques, and \(r_{\textit{bc}}\) penalizes early termination of episode caused by base collision.

\subsection{Reward Relaxation} \label{sec:2-3}
After presenting the effective learning method for dynamic jumping by combing TO and RL, we aim to introduce a method to gain the ability to landing on uneven terrains adaptively. Unlike the reference motion for flat ground that four feet touch down simultaneously, some feet are likely to touch down earlier than others when landing on uneven terrains, which causes the dynamically unbalanced contact wrench that tilts the base pose, and at last causes the failure of landing. Another problem introduced by rough terrains is the conflict between the joint and base reference information. To land safely, the actual joint angles must deviate from the optimized joint angles for terrain adaptation. Inspired by whole body control (WBC) that optimizes joint torques according to task priority~\cite{bellicoso_2018}, it is most important to maintain the balance of robot body, i.e., the base pose, before minimizing the tracking error of joint positions. 

We propose a reward relaxation on the tracking of joint references for a compliant jumping controller. The tracking reward function in Equation~\eqref{eq:tracking_reward} can be regarded as a soft constraint, as the gradient vanishes to zero when the reward value reaches its maximum. We create a WBC-like behavior for the convergence priority of different tracking reward terms by decreasing the gradient of the joint tracking reward function by modifying \(\sigma\) after the touch-down moment rolled out from TO. This would prioritize the tracking of base pose because of its steeper gradient. As the base tracking reward converges, its gradient decreases, which naturally shifts the optimization focus toward refining joint tracking objectives. The value of \(\sigma\) in stance and flight phase (S\&F) and in landing phase(L) is shown in Table~\ref{tab:reward_functions}. 
\begin{table}[h]
\centering
\caption{Reward Functions}
\label{tab:reward_functions}
\scalebox{1.05}{
\begin{tabular}{lcl}
\toprule
Reward & Weight & Parameter \\ \midrule
Base position tracking & 5 & $\sigma=0.02$ \\
Base rotation tracking & 2 & $\sigma=0.01$ \\
Joint position tracking (S\&F) & 2 & $\sigma=0.2$ \\
Joint position tracking (L) & 2 & $\sigma=2$ \\
Action rate & $-0.01$ &  \\
Action smoothness & $-0.02$ &  \\
Joint power & $-2e-5$ &  \\  
Joint torque & $-1e-5$ &  \\ 
Base collision & $-10$ &  \\ \bottomrule
\end{tabular}}
\end{table}

\section{Results}
In this section, we present implementation details of motion planning and policy training, as well as experimental validations using \textit{Unitree A1}.

\subsection{Implementation Details}
\subsubsection{Motion Planning}
For jumping TO setups, the homing base height is 0.25 m, and the initial joint angles are [0, 0.896, -1.791]~rad for all legs. If not specified, the shape of feet polygon is not changed during the whole jumping period. The projection of the distance between each foot and the CoM of the body on the local \textit{x}-\textit{y} plane is [0.181, 0.131]~m at homing pose. The number of knots for both stance phase \(N_s\) and flight phase \(N_f\) is 100. The nonlinear optimization problem is solved using CasADi~\cite{andersson_casadi} in Python with the IPOPT solver~\cite{Wchter2006}. 

\subsubsection{Policy Training}
Only one reference jumping motion is tracked for a single training process. The trajectory planner rolls out a reference motion at the start of the training process. The reference trajectories are then linearly interpolated with a constant timestep that aligns with the policy frequency. The training is conducted on a discrete obstacles terrain with parameters shown in Table~\ref{tab:terrain_parameters}. 
During the training, we randomize the friction coefficient of the terrain and add perturbed noise to observations. The policies are trained using NVIDIA's Isaac Gym~\cite{makoviychuk2021isaacgym} and the legged\_gym massively parallel RL framework based on PPO~\cite{rudin2022leggedgym}. Both the actor and critic networks are a 4-layer MLPs with hidden layer sizes (512, 256, 128) and ELU activations. The training hyperparameters are listed in Table \ref{tab:ppo_hyperparameters}. We set the maximum episode length as 3 seconds, and train policies with a parallelization of 4096 environments for 15,000 iterations, in which 24 steps are taken from each environment.
\begin{table}[h]
\centering
\caption{Terrain Parameters}
\label{tab:terrain_parameters}
\scalebox{1.05}{
\begin{tabular}{lc|lc}
\toprule
Name & Value & Name & Value \\ \midrule
Max height & 0.1 m& Num. of obstacles & 1500 \\
Mini size & 0.4 m& Platform size & 5 m\\
Max size & 2 m&  &  \\ \bottomrule
\end{tabular}}
\end{table}

\begin{table}[h]
\centering
\caption{PPO Hyperparameters}
\label{tab:ppo_hyperparameters}
\scalebox{1.05}{
\begin{tabular}{lc|lc}
\toprule
Name & Value & Name & Value \\ \midrule
Value loss coeff. & 1 & Learning rate & Adaptive \\
Entropy coeff. & 0.01 & Discount factor & 0.99 \\
Clip range & 0.2 & GAE parameter & 0.95 \\
Num. of epochs & 5 & Max gradient norm. & 1 \\
Mini batches & 4 &  &  \\ \bottomrule
\end{tabular}}
\end{table}

\subsubsection{Deployment}
Similar to the training stage, the reference trajectories are first generated via TO at deployment startup, then interpolated to match the policy's control frequency for real-time execution. Both the TO module and the RL policy are run on a laptop equipped with an NVIDIA GeForce RTX 3060 GPU and Intel Core i7-12700H CPU. The control policy runs at 50 Hz, and the lower level PD controller computes torques at 1000 Hz.

\subsection{Hardware Experiment}
Experiments on the real A1 robot aim to show accurate dynamic jumping and compliant landing skills benefiting from our proposed method. In the following experiments, policies were trained using three methods for comparison and ablation studies: 1) a baseline policy trained on flat ground without reward relaxation, 2) an ablated policy trained on rough terrains without reward relaxation and 3) a policy trained on rough terrains using our complete proposed method. Except for the differences mentioned above, the three policies were trained using the same setups for a 0.8 m forward jump, which were deployed to A1 robot without any fine-tuning. We tuned PD gains for each policy to optimize its performance, and all three policies shared the identical \(\mathbf{K}_P=20\mathbf{I}\), and the differential gains are, respectively, \(\mathbf{K}_{D\_baseline}=0.8\mathbf{I}\), \(\mathbf{K}_{D\_ablation}=\mathbf{I}\) and \(\mathbf{K}_{D\_ours}=0.95\mathbf{I}\), where \(\mathbf{I}\) denotes identity matrix.

\subsubsection{Jumping on Flat Ground}
\begin{figure}[t]
  \centering 
  \begin{subfigure}[t]{0.48\columnwidth}
    \includegraphics[width=\linewidth]{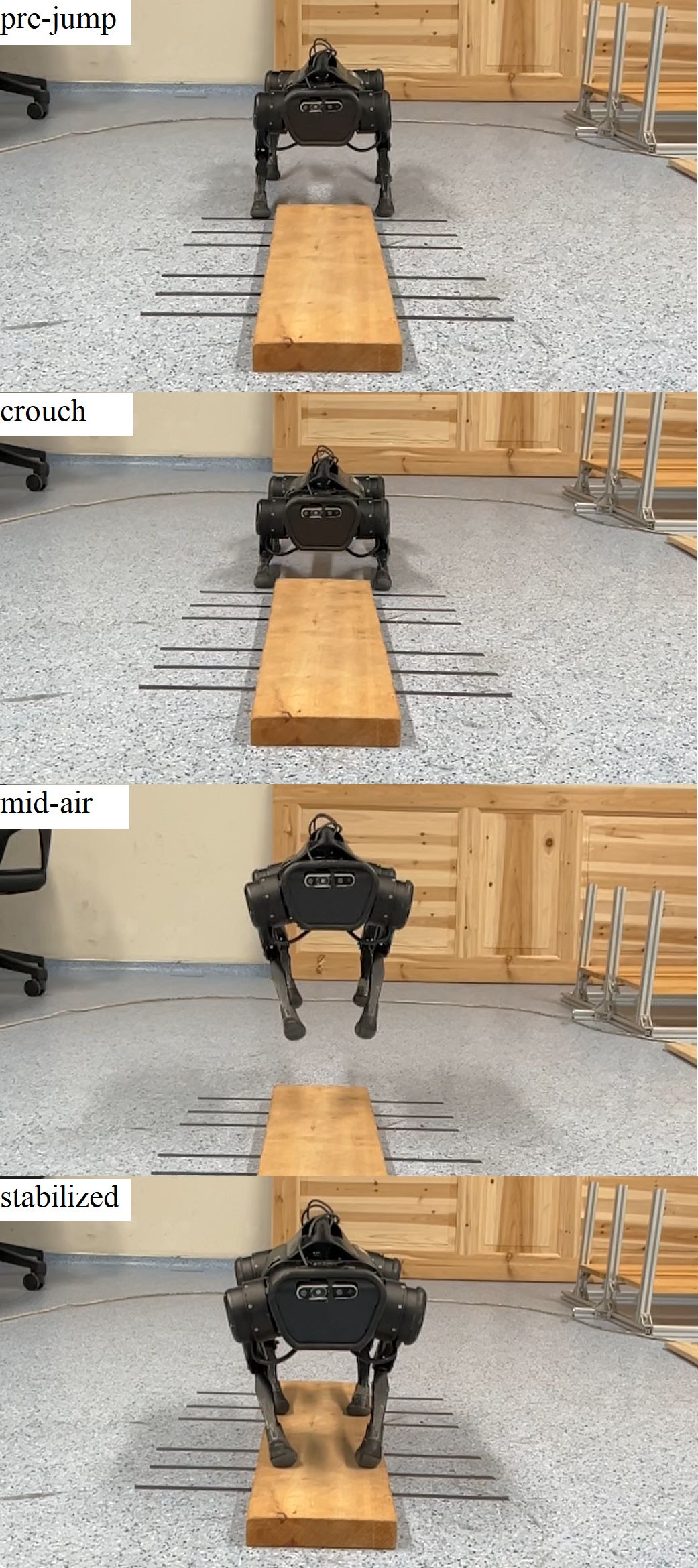}
    \subcaption{Narrow retract}
    \label{fig:narrow_retract}
  \end{subfigure}
  \hfill 
  \begin{subfigure}[t]{0.48\columnwidth}
    \includegraphics[width=\linewidth]{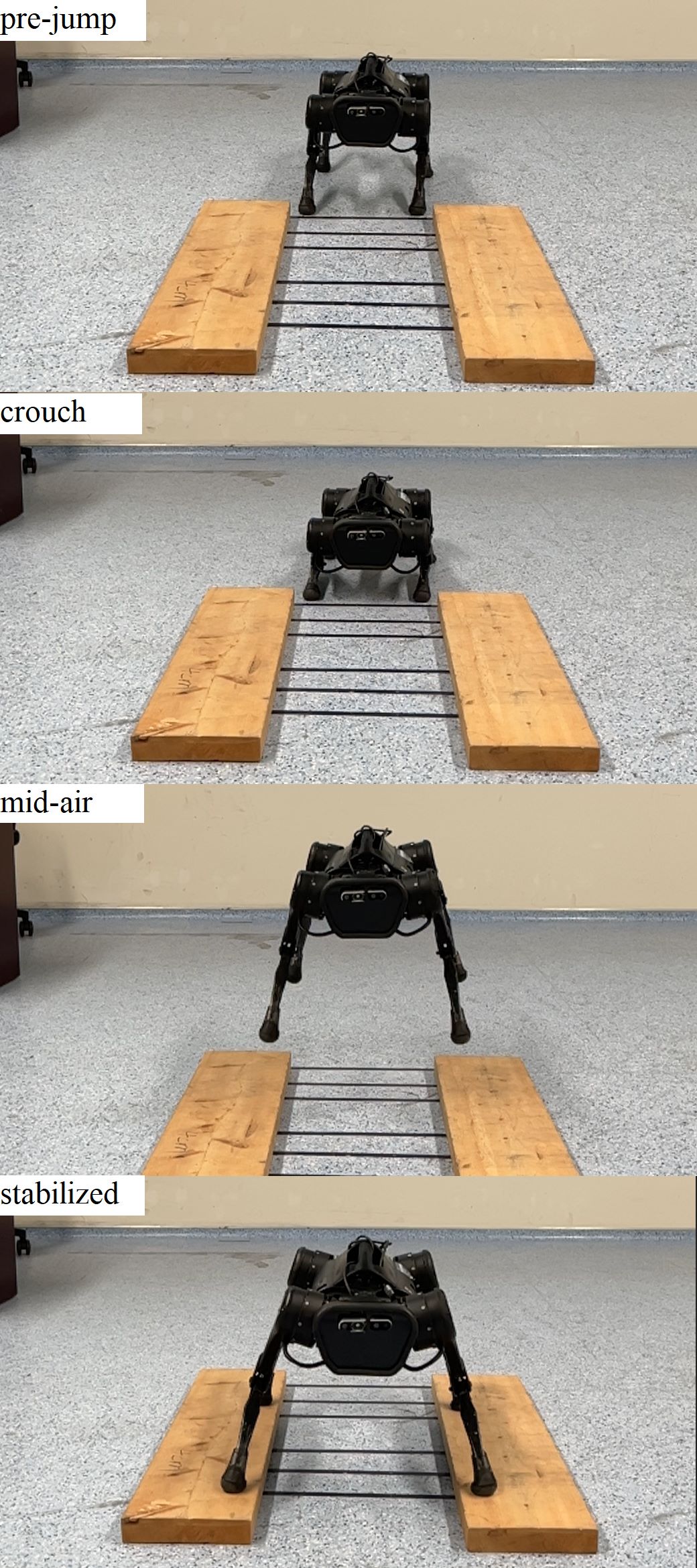}
    \subcaption{Split expand}
    \label{fig:split_expand}
  \end{subfigure}
  \caption{Experiment snapshots of (a) narrow landing and (b) wide landing}
  \label{fig:photo_exp1}
\end{figure}

This experiment aims to demonstrate accurate tracking of the reference motion benefiting from our combined learning framework. The robot was tasked to separately jump onto a narrow area and a wide area with a gap in the middle resembling sparse landing scenarios, which were set up using 6~cm high wooden planks. These tasks are critical for an accurate landing of the four feet, as shown in Fig.~\ref{fig:photo_exp1}. It is obvious to conclude that landing on both target areas cannot be successful with the homing feet polygon. In order to achieve this, We trained the narrow and wide jumping policies using the baseline setups, which separately mimic two modified jumping motions rolled out from the planner. We changed the shape of the feet polygon after the take-off of each task. For the narrow landing task, we shrank the width of the polygon by 12~cm, (i.e., 6~cm for each leg), and move the two rear feet forward by 6~cm. For the other task, the polygon was widened by 16~cm and the two front feet were put forward by 6~cm. 

\begin{figure}[t]
    \centering
    \includegraphics[width=1\linewidth]{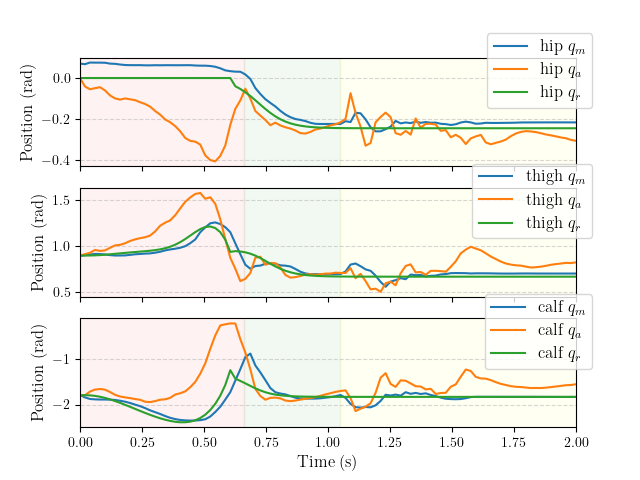}
    \caption{Joint positions of the hind left leg performing narrow jumping. \(q_m\), \(q_a\) and \(q_r\) respectively denote measured joint position, desired angle given by policy and reference joint angle from trajectory planner. The shallow red, green, and yellow stripe zones separately cover the real stance, flight, and landing motions.}
    \label{fig:exp1_1}
\end{figure}

\begin{figure}[h!]
    \centering
    \includegraphics[width=1\linewidth]{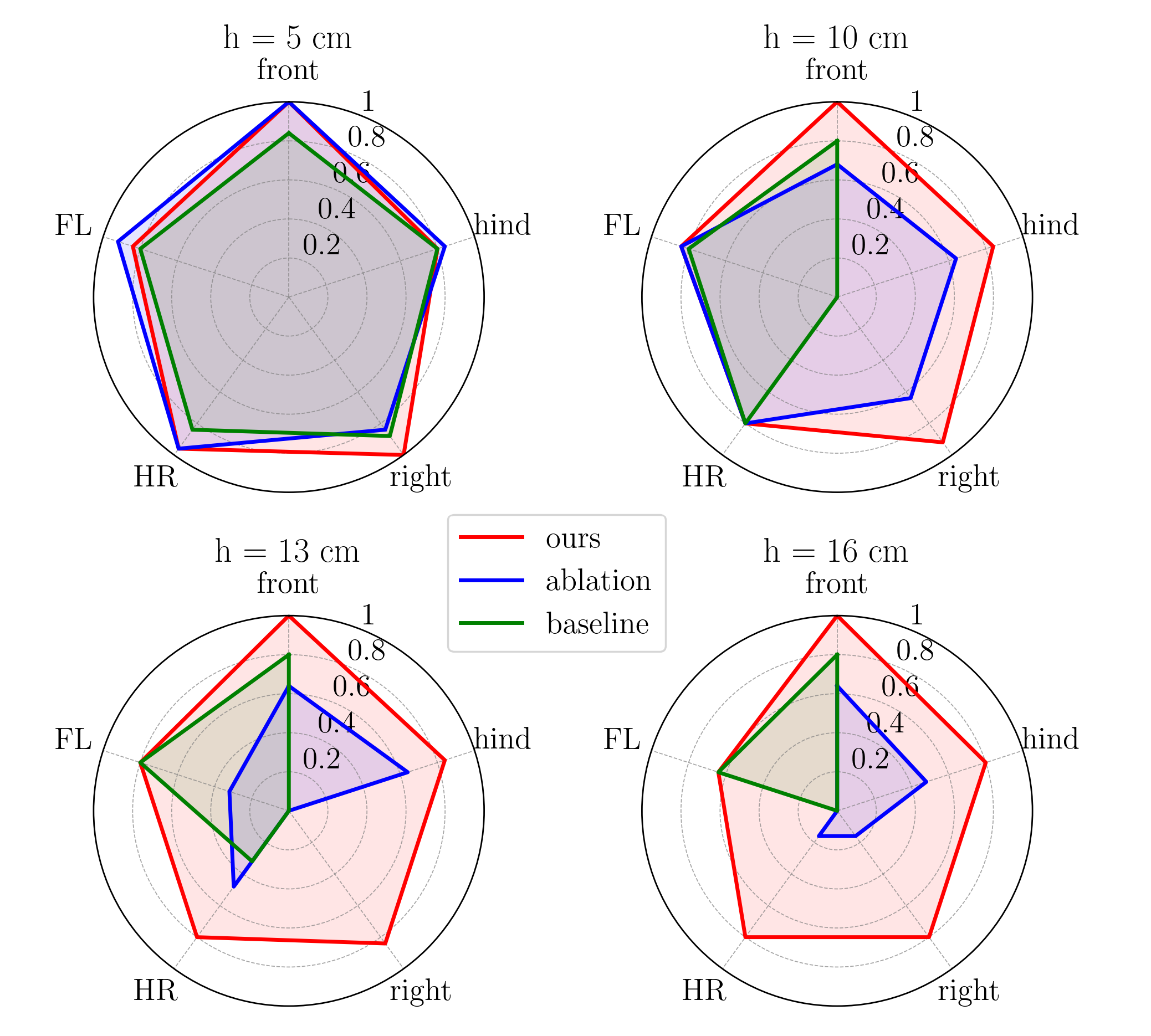}
    \caption{Success rates for 5 scenarios over each of 4 difficulty levels. The radial axis denotes success rate, while the angular axis represents the name of scenarios. The front, hind, right, HR, and FL stand for, respectively, landing scenarios where the front feet, the hind feet, the right-hand-side feet, the hind-right foot and the front-left foot landed on the obstacles.}
    \label{fig:exp2_1}
\end{figure}

The snapshots of the experiment are illustrated in Fig.~\ref{fig:narrow_retract} and Fig.~\ref{fig:split_expand} for jumping onto a narrow plank and split planks respectively, which demonstrate the change in the feet polygon after take-off in order to match the geometry of the landing area. More straightforward tracking performance can be seen from joint space, for which the hind left leg for the narrow retract landing is shown in Fig.~\ref{fig:exp1_1}. This signifies the accurate joint-level tracking achieved by our learning framework that combines TO and RL in dynamic jumping motion and therefore could benefit landing on sparse terrains when incorporated with exteroceptive perception.  

\subsubsection{Jumping onto Rough Terrains}

\begin{figure*}[t]
  \centering
  \includegraphics[width=1\textwidth]{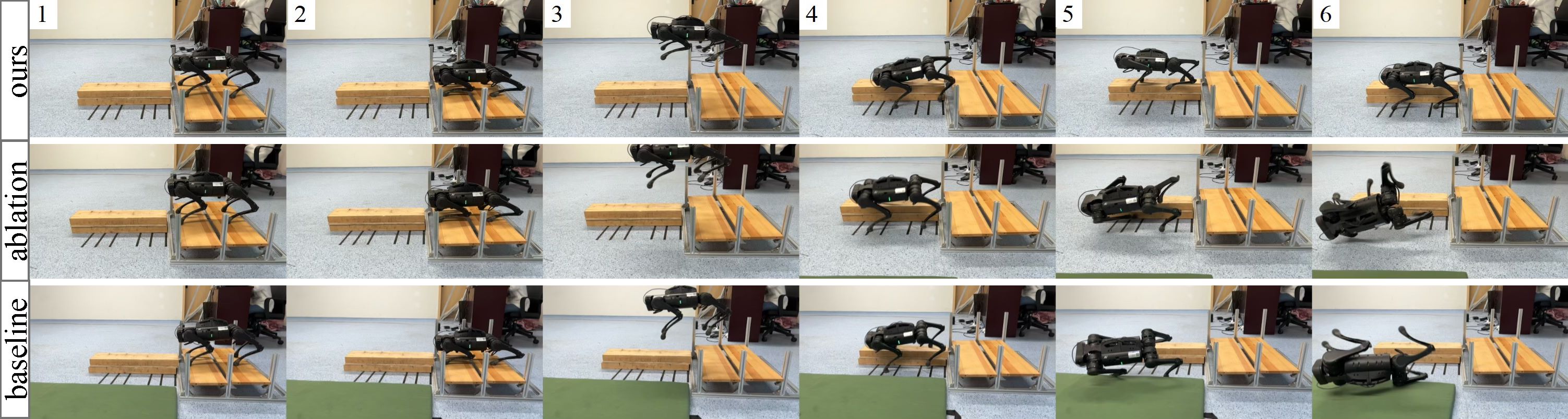} 
  \caption{Snapshots of jumping onto 13 cm high obstacles placed at the landing location of the right-hand-side feet.}
  \label{fig:photo_exp2}
\end{figure*}

\begin{figure*}[h]
  \centering
  \includegraphics[width=1\textwidth]{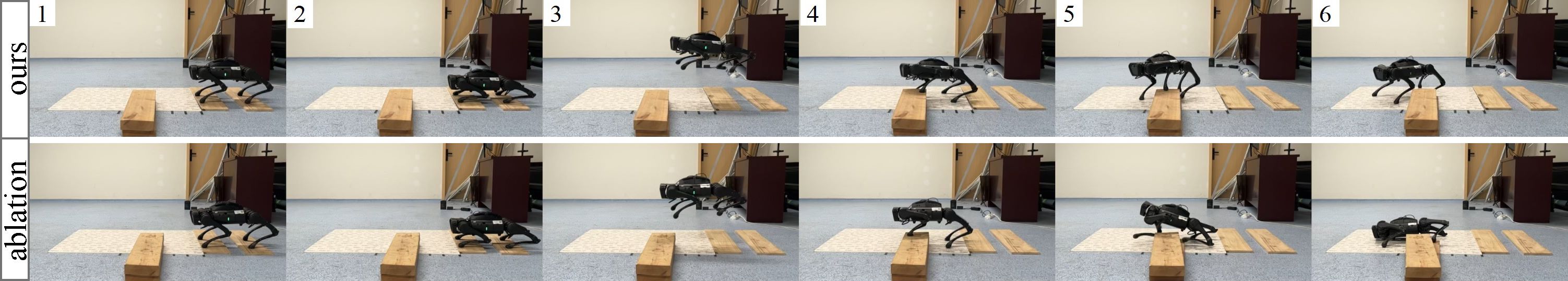} 
  \caption{Snapshots of jumping onto 10 cm high obstacles placed on a slippery white board at the landing position of the front-left foot.}
  \label{fig:photo_exp3}
\end{figure*}

\begin{figure*}[h]
  \centering
  \includegraphics[width=1\textwidth]{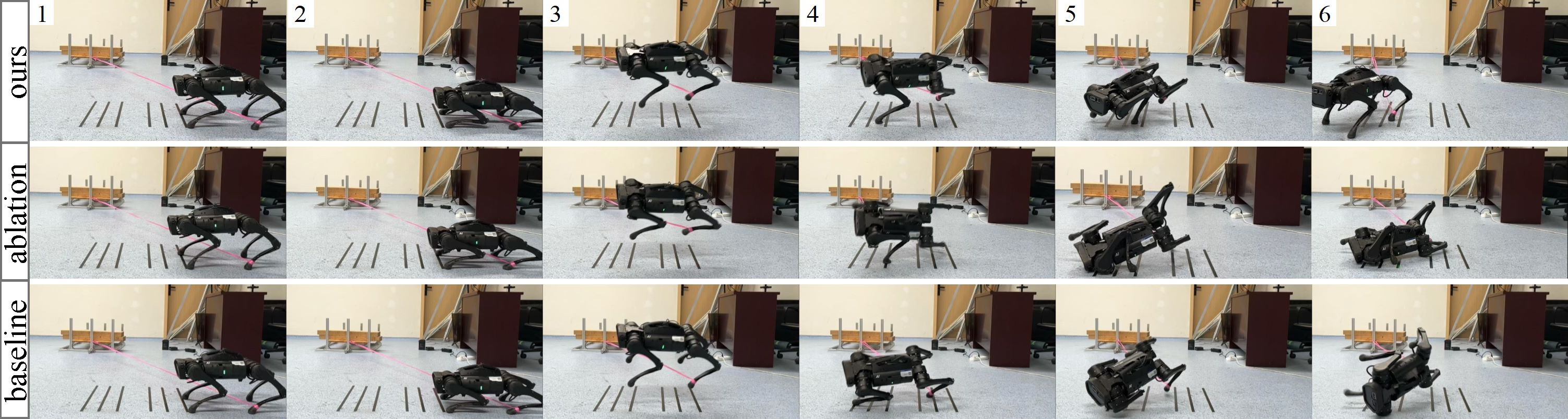} 
  \caption{Snapshots of disturbed jumping experiment.}
  \label{fig:photo_exp4}
\end{figure*}

In our second experiment, we validate the compliant landing ability in extensive scenarios. We created five different uneven landing scenarios using wooden planks as unknown landing obstacles that cause early feet touch-down. In each scenario, we placed the obstacles on the landing spots of different feet, i.e., the front feet, the hind feet, the right-hand-side feet, the front-left foot and the hind-right foot. Each scenario contains four levels of difficulty with increasing obstacle height, including 5~cm, 10~cm, 13~cm and 16~cm. We conducted this experiment using the three policies for 5 times per terrain per difficulty level and per policy. In order to enable the robot to land on the planks with the desired feet, we let the robot jump from an elevated platform with height of 10 cm for the scenarios requiring the rear feet to land on the obstacles. The snapshots of the experiment jumping onto 13 cm obstacles which are under the right-hand-side feet are shown in Fig.~\ref{fig:photo_exp2}. 
We classified the results into three categories for the evaluation of compliant landing capability: 1) success (S) for a stable landing and a final stop of the corresponding feet on the obstacles, 2) weak success (WS) if the feet hopped off the obstacles and finally stabilized, and 3) fail (F) if the robot either knocked its body, exceeded joint torque limits, or failed to stabilize itself within 1~second after touch-down. We assigned a score for each category, i.e., 1 for S, 0.8 for WS, and 0 for F. After that, we computed the success rates and the results are presented in Fig.~\ref{fig:exp2_1}. 

From Fig~\ref{fig:exp2_1} it can be seen that the three policies performed almost equally on 5 cm obstacles. As the difficulty level increased, significant performance degradation was observed on both the ablated and baseline methods for almost all scenarios, while a minor drop in success rates is shown with the proposed method. Note that the maximum height of obstacles during the training for both the ablated and the proposed policies was 10 cm. The ablated method still retained part of the safe landing capability on 10 cm high obstacles, but it degraded significantly on 13 cm and 16 cm high obstacles. On the contrary, the proposed method generalized adaptive landing ability to 16 cm successfully, which signifies the effectiveness of reward relaxation. 

\begin{figure}[t]
    \centering
    \includegraphics[width=1\linewidth]{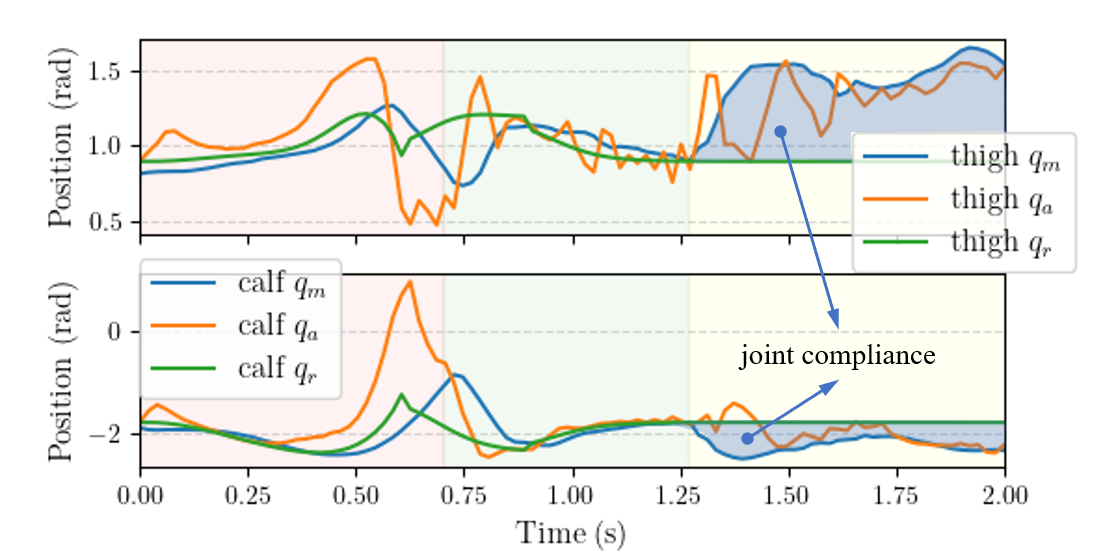}
    \caption{FR thigh and calf joint positions jumping onto 13 cm obstacles placed under the right-hand-side feet with the proposed method}
    \label{fig:exp2_2}
\end{figure}

\begin{figure}[t]
    \centering
    \includegraphics[width=1\linewidth]{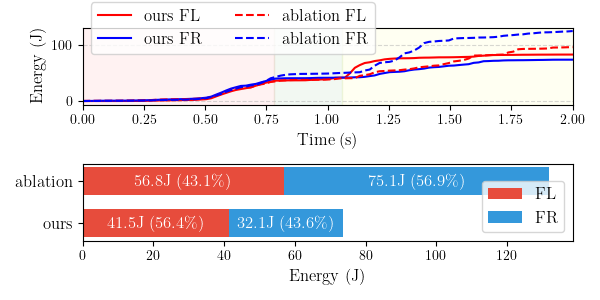}
    \caption{Energy consumed jumping onto 10 cm obstacles placed under the right-hand-side feet.}
    \label{fig:exp2_3}
\end{figure}

This huge improvement of success rates benefits from the learned compliant joint behavior, which helps to keep the body's nominal pose on rough terrains as shown in the first row of Fig.~\ref{fig:photo_exp2}. A more detailed representation of this capability is shown in Fig.~\ref{fig:exp2_2}. The front right (FR) thigh and calf joints adaptively deviated from the reference trajectories after touch-down (the shallow blue area) as the corresponding leg was standing high. This behavior can be interpreted from the perspective of energy. We computed the cumulative energy consumed with time for the front left (FL) leg that landed on the ground and the FR leg that landed on the terrain, and the total energy consumed during the landing recovery period for the two legs, both with the proposed and the ablated methods on 10 cm obstacles where neither policy failed, as seen in the first and the second row in Fig.~\ref{fig:exp2_3} respectively. Both of the policies consumed approximately equal energy for taking off. For landing recovery, less energy was consumed by the FR leg (43.6\%) than by the FL leg (56.4\%) with the proposed method. During this process, the FR and the FL legs separately resembled springs with lower and higher stiffness adapted to the terrain. In contrast, the two legs had the same stiffness for the abalted method, thus more energy was consumed by the FR leg due to early touch-down. Moreover, Fig.~\ref{fig:exp2_3} also signifies that the proposed method is more energy efficient.

\subsubsection{Jumping onto Rough and Slippery Terrains}

\begin{figure}[t]
    \centering
    \includegraphics[width=1\linewidth]{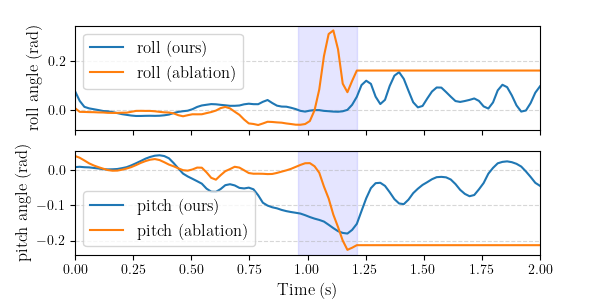}
    \caption{Robot roll and pitch rotation jumping onto 10 cm obstacles placed under the front-left foot with slippery ground with the proposed and the ablated methods. The shallow purple zone highlights the landing recovery period.}
    \label{fig:exp3_1}
\end{figure}
To further explore the limit of our reward relaxation strategy, we modified the previous experiment's setups by making the ground more slippery using a white board with soap. This thereby imposes stricter demands on compliant joint control and dynamic balance against unforeseen environments. We evaluated only the proposed and ablated methods, as such challenging scenario is unfair for the baseline method that was trained on flat ground. We let the robot jump onto 10 cm high planks that were placed on the white board at the landing position of the front left foot. The experiment snapshots are shown in Fig.~\ref{fig:photo_exp3}. As seen in the figure, during the landing recovery period, the proposed method showed better dynamic balance with minor feet slip, while the robot with ablated method lost balance due to aggressive slip and ended with exceeding torque limits. As seen in Fig~\ref{fig:exp3_1}, the slip with ablated method is reflected by the rapid change in roll and pitch angles, while the body rotation with the proposed method is much smaller. This validates that the learned compliant joint control with the proposed method is able to adapt to changes in the environment.  

\subsubsection{Jumping with Unknown Disturbances}

\begin{figure}[t]
    \centering
    \includegraphics[width=1\linewidth]{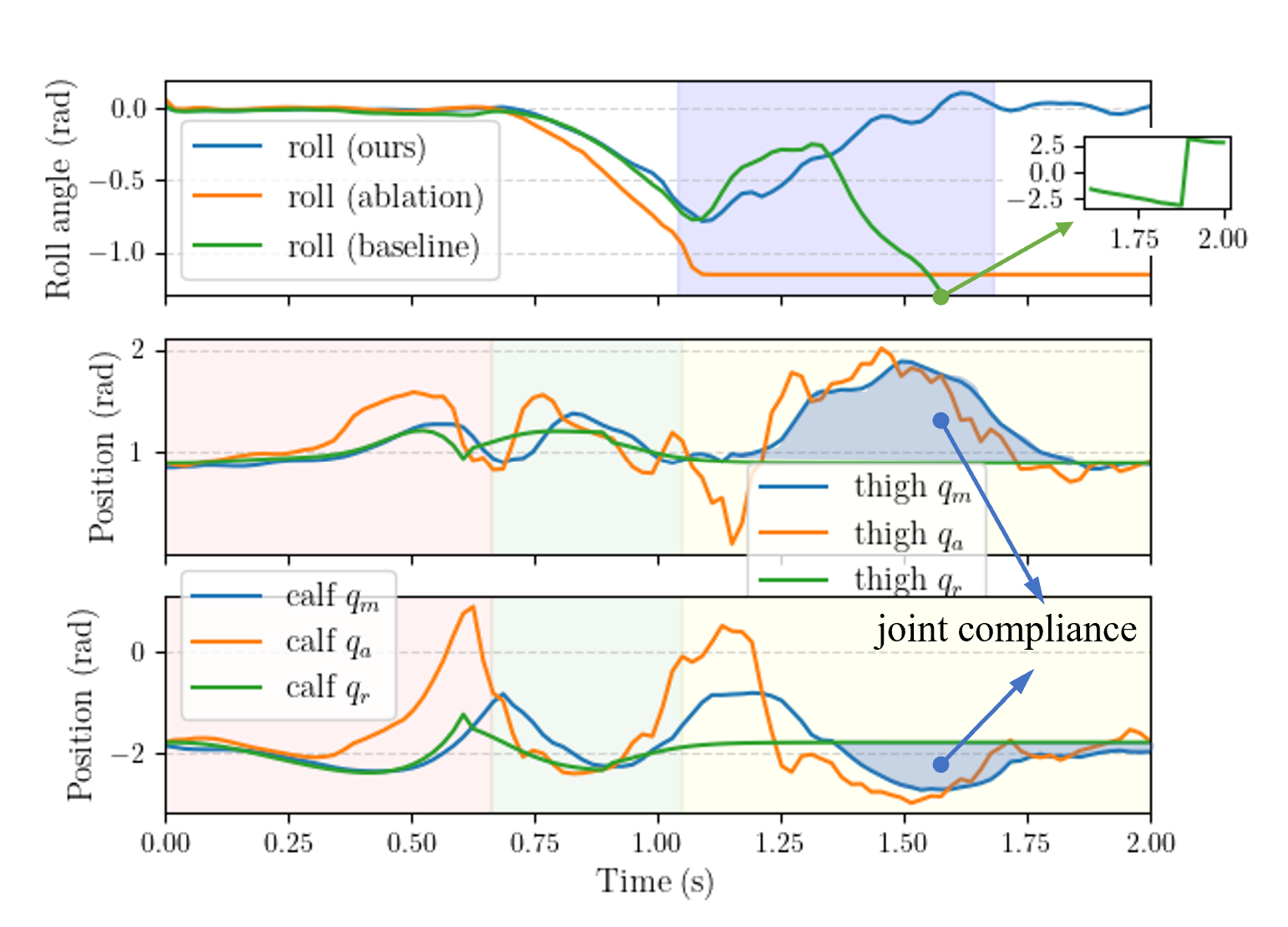}
    \caption{Robot states of the disturbed jumping experiment. The first row represent body roll angle for the proposed, the ablated and the baseline policy, and the second and third rows show respectively the thigh and calf positions of the hind-left leg for the proposed method. The shallow purple zone highlights the landing recovery period.}
    \label{fig:exp4_1}
\end{figure}
In this experiment we applied external disturbances on the robot during the whole jumping period, using a resistance band with one end tied on the hind left ankle of the A1 robot and the other end fixed to an object located on the right-hand side of the robot. This disturbance affected both joint tracking of the reference joint motion and the dynamic balance of the body pose. We conducted this experiment and performed flat ground jumping task using the proposed, the ablated and the baseline policies. The experiment snapshots are demonstrated in Fig.~\ref{fig:photo_exp4}, which shows that only the proposed method succeeded in this experiment. 
As seen in the first row of Fig.~\ref{fig:exp4_1}, the disturbance caused base rotation in roll angle during flight phase, for all trials using the three policies. During this process, the proposed policy survived with a maximum roll angle of 0.78~rad (or about \(45^\circ\)), while the ablated and baseline methods ended with exceeding torque limits and roll-over, respectively.
The success of the proposed method can be interpreted by adaptive joint behavior.
As seen in the first row of Fig.~\ref{fig:photo_exp4}, at the descending period (the fourth snapshot), the proposed method mitigated the impact of the resistance band by stretching its hind-left leg towards the other end of the band to shorten the band length and decrease elastic force while extending its front-left leg to make it touch down first to recover from the disturbed state. After the robot stabilized, these joints recovered to their homing pose. This adaptive joint behavior that deviates from the impractical reference joint trajectories is demonstrated in the second and the third rows of Fig.~\ref{fig:exp4_1}.   
This verifies the robustness and adaptability of the proposed method.

\section{Conclusion}
Quadrupedal dynamic jumping with safe landing skills on rough terrains is presented in this work. This challenging task is achieved by integrating the best of both trajectory optimization and reinforcement learning, as well as an intuitive reward relaxation strategy. Extensive real-world experiments validate both accurate command-tracking ability and a remarkable survival rate in unknown environments with disturbances. Future work includes synthesizing a faster trajectory optimization and an adaptive reward relaxation scheme.


\bibliographystyle{IEEEtran}
\bibliography{ref}

\addtolength{\textheight}{-12cm}   

\end{document}